\documentclass{article}

\usepackage{arxiv}

\usepackage[utf8]{inputenc} 
\usepackage[T1]{fontenc}    
\usepackage{hyperref}       
\usepackage{url}            
\usepackage{booktabs}       
\usepackage{amsfonts}       
\usepackage{nicefrac}       
\usepackage{microtype}      
\usepackage{lipsum}
\usepackage{amsmath}
\usepackage{graphicx}

\title{Tag and Correct: Question aware Open Information Extraction with Two-stage
Decoding}

\author{
 Martin Kuo\\
  Microsoft\\
  Beijing, China \\
  \texttt{martintw.kuo@gmail.com} \\
   \And
 Yaobo Liang\\
  Microsoft\\
  Beijing, China \\
  \texttt{yalia@microsoft.com} \\
  \And
 Lei Ji\\
  Microsoft\\
  Beijing, China \\
  \texttt{leiji@microsoft.com} \\
  \And
 Nan Duan\\
  Microsoft\\
  Beijing, China \\
  \texttt{nanduan@microsoft.com} \\
  \And
 Linjun Shou\\
  Microsoft\\
  Beijing, China \\
  \texttt{lisho@microsoft.com} \\
  \And
 Ming Gong\\
  Microsoft\\
  Beijing, China \\
  \texttt{migon@microsoft.com} \\
  \And
 Peng Chen\\
  Microsoft\\
  Beijing, China \\
  \texttt{peche@microsoft.com} \\
}

\begin{document}
\maketitle

\begin{abstract}
Question Aware Open Information Extraction (Question aware Open IE) takes question and passage as inputs, outputting an answer tuple which contains a subject, a predicate, and one or more arguments. Each field of answer is a natural language word sequence and is extracted from the passage. The semi-structured answer has two advantages which are more readable and falsifiable compared to span answer. There are two approaches to solve this problem. One is an extractive method which extracts candidate answers from the passage with the Open IE model, and ranks them by matching with questions. It fully uses the passage information at the extraction step, but the extraction is independent to the question. The other one is the generative method which uses a sequence to sequence model to generate answers directly. It combines the question and passage as input at the same time, but it generates the answer from scratch, which does not use the facts that most of the answer words come from in the passage. To guide the generation by passage, we present a two-stage decoding model which contains a tagging decoder and a correction decoder. At the first stage, the tagging decoder will tag keywords from the passage. At the second stage, the correction decoder will generate answers based on tagged keywords. Our model could be trained end-to-end although it has two stages. Compared to previous generative models, we generate better answers by generating coarse to fine. We evaluate our model on WebAssertions ~\cite{yan2018assertion} which is a Question aware Open IE dataset. Our model achieves a BLEU score of 59.32, which is better than previous generative methods.
\end{abstract}


\section{Introduction}

Question aware Open Information Extraction (Question aware Open IE) system takes question and passage as inputs and extracts a semi-structure answer in tuple format from passage which can answer the question. Question aware Open IE is both an Open IE task and a question answering task. From an Open IE view, the Open IE system extracts all possible tuples. For example, in Table \ref{table_openIE}, Open IE system aims to extract four answer tuples from the passage, which are independent of any questions. A question aware Open IE only extracts one answer tuple which can answer the question. From a question answering view, the answer of the search engine is a passage; the answer for Machine Reading Comprehension tasks like SQuAD~\cite{rajpurkar2016squad}, TriviaQA ~\cite{joshi2017triviaqa} and NewsQA ~\cite{trischler2016newsqa} is a span from the passage; the answer of MS MACRO ~\cite{nguyen2016msmacro} is a generated sentence. Different to them, the answer for a question aware Open IE is a semi-structure tuple which is shorter than the passage and longer than the span. It has a semantic role for each part which is easier for understanding for downstream task. \\
\begin{table}[t!]
	\begin{center}
		\begin{tabular}{p{0.7in}|p{2.3in}}
        	\hline \textbf{Question} & how many albums has the doors sold \\
            \hline \textbf{Passage} & 
              although the doors' active career ended in 1973 , their popularity has persisted. according to the riaa, they have sold over 100 million records worldwide, making them one of the best-selling bands of all time.
              \\
              \hline 
            
            \textbf{Open IE Result}
            
            & 
            \begin{minipage}{0.7\columnwidth}
(the doors active career; ended; in 1973) \\
(their popularity; has persisted; although the doors active career ended in 1973) \\
(they; making; them one of the best-selling bands of all time) \\
(they; have sold; 100 million records worldwide) \\
            \end{minipage}\\
            \hline
            
            \textbf{Answer (Question aware Open IE)}&
            (they; have sold; 100 million records worldwide)
            \\
            \hline
        \end{tabular}
        \caption{Example of Open IE and Question aware Open IE. The Open IE Result and Answer has same format, which is (subject; predicate; arguments), there could be more than one argument. The column "Open IE Result" is tuples extracted by Open IE tools from passage independent with question. The "Answer" is extracted from passage and could answer the question.}
        \label{table_openIE}
	\end{center}
\end{table}
The current solution for a question aware Open IE has two approaches, the extractive method and the generative method. The extractive method extracts all possible answer tuples as candidates from the passage, independent of the question, by Open IE models. It then ranks all the candidates by a matching model between candidate and question. The coverage of the extraction step is crucial for the final performance because it is a twostep method and the extraction step is independent of the question. Since the first step is extraction, most of the words in the result will come from the passage.

The generative method concatenates the question and passage as input, and then generates the answer tuple as a concatenated sequence or generates each field one by one. The generative method uses the question and passage at the same stage and does not rely on an extraction model, so it has better interaction between question and passage. But while it removes the extraction step, it does not use the facts that most of the answer word is from passage any more.

To better use passage information in generation, we propose a two-stage decoder model for this task which can have more interaction between question and passage by containing a tagging decoder and a correction decoder. At the first stage, the tagging decoder will tag the words which may be useful for answer generation. The output of this step could form a coarse answer. At the second stage, the correction decoder generates a new answer with a step by step decoder. A correction decoder can reorder and add new words to output a fluency answer. Then we joint train two decoders at the same time.

We evaluated our model on the WebAssertion dataset ~\cite{yan2018assertion}. Our model achieves a 59.32 BLEU score which is better than previously generative methods.

\begin{table*}[t!]
	\centering
    \begin{tabular}{p{1.1in}|p{5.5in}}
    \hline \bf Question & where is smallville filmed \\
    \hline \bf Passage & smallville was primarily filmed in and around vancouver , british columbia , with local businesses and buildings substituting for smallville locations . \\
    \hline \bf Answer & smallville; was filmed; in british columbia; with local businesses \\
    \hline \bf Tagging Label & $smallville_{S-B}\; was_{P-B}\; primarily_{O}\; filmed_{P-B}\; in_{A_0-B}\; and_{O}\; around_{O}\; vancouver_{O}\; ,_{O}\;$ 
    $british_{A_0-B}\; columbia_{A_0-I}\; ,_{O}\; with_{A_1-B}\; local_{A_1-I}\; businesses_{A_1-I}\; and_{O}\; buildings_{O}\;$  
    $substituting_{O}\; for_{O}\; smallville_{O}\; locations_{O}\; ._{O}$\\
    \hline \bf Tagging Result &smallville, was primarily filmed, vancouver, british columbia, with local businesses \\
    \hline \bf Correction Result & smallville, was filmed, in vancouver, british columbia with local businesses \\
	\hline
    \end{tabular}
	\caption{Example of tagging label and model output of tagging decoder and correction decoder. The tagging label is created from answer since original dataset only has tuple format answer.}
	\label{table_tagging}
\end{table*}

\section{Related Work}
\textbf{Open Information Extraction (Open IE)} ~\cite{banko2007open,etzioni2008open} aims to extract all (subject, predicate, arguments) tuples from a sentence. To solve this challenge, TextRunner ~\cite{banko2007open} and WOE ~\cite{wu2010WOE} use a self-supervised approach. Then many of the methods use a rule based approach like ReVerb ~\cite{fader2011ReVerb}, OLLIE ~\cite{schmitz2012OLLIE}, 
KrakeN ~\cite{akbik2012kraken}, ClauseIE ~\cite{del2013clausie} PropS ~\cite{stanovsky2016props}. Open IE4\footnote{https://github.com/dair-iitd/OpenIE-standalone} extracts tuples from Semantic Role Labeling structures. Stanford Open Information Extraction ~\cite{angeli2015leveraging} uses natural logic inference to extract a shorter argument.
Recently, Stanovsky et al. ~\cite{stanovsky2018supOpenIE} have proposed a supervised method for Open IE by formulating it as a sequential labeling task. Compared to Open IE, our task has an additional question, so our tagging decoder needs to contain an interactive layer between question and passage. Our tagging decoder is similar to ~\cite{stanovsky2018supOpenIE}, since they have the same output and trained by supervised learning. However, we have an additional correction decoder to improve answer quality and can handle an answer field that is not a span.

Current \textbf{Machine Reading Comprehension (MRC)} like SQuAD ~\cite{rajpurkar2016squad}, TriviaQA ~\cite{joshi2017triviaqa} and NewsQA ~\cite{trischler2016newsqa} focus on selecting a span from passage as the answer. Most MRC models ~\cite{wang2016pointerNetWork,wang2017rnet,yu2018qanet} generate answers by predicting the start and end points of a span. The MS MACRO ~\cite{nguyen2016msmacro} dataset needs to generate a sequence which is not a span of a passage. Tan et al. ~\cite{tan2017snet} solve it by selecting a span from the passage at first, then generates an answer based on the question, passage and selected span. Similar to Tan et al. ~\cite{tan2017snet}, we also used the idea of coarse to fine generation,
but the answer of our task is not a span or sentence. Our answer has structure and each field has a semantic role. The arguments have dynamic length. Each field of it does not have to be a span, although most of the words are from a passage. because of this we use a sequential labeling method to tag each word in a passage instead of predicting the start and end point of a span. The two stages of our model could be jointly trained.

For \textbf{Question aware Open IE}, Yan et al. ~\cite{yan2018assertion} propose two methods, an extractive method and a generative method. The extractive method extracts all answer tuples from the passage first, and ranks them with a matching model between answer candidates and questions. The generative model takes the concatenation of question and passage as input, and generates the representation of each answer field at first, and then generates each field based on its representation.

\section{Our Approach}

In this section, we formulate the Question aware Open IE problem and briefly introduce our model. Then we separately introduce each part of our model including the encoder, tagging decoder, and correction decoder.

\subsection{Problem Formulation}

The Question aware Open IE is a task which given a question containing $n$ words $Q=\{q_1,q_2,...,q_n\}$ and a passage containing $m$ words $P=\{p_1,p_2,...,p_m\}$, and output a semi-structured answer which can answer the question based on passage. The answer consists of a subject, a predicate, and one or more arguments. We represent the answer as $(subject, predicate, argument_1, ..., argument_k)$, $k\geq1$. Each answer field is a natural language word sequence.

\subsection{Model Overview}
\begin{figure*}[t!]
\centering
\includegraphics[width=0.8\linewidth]{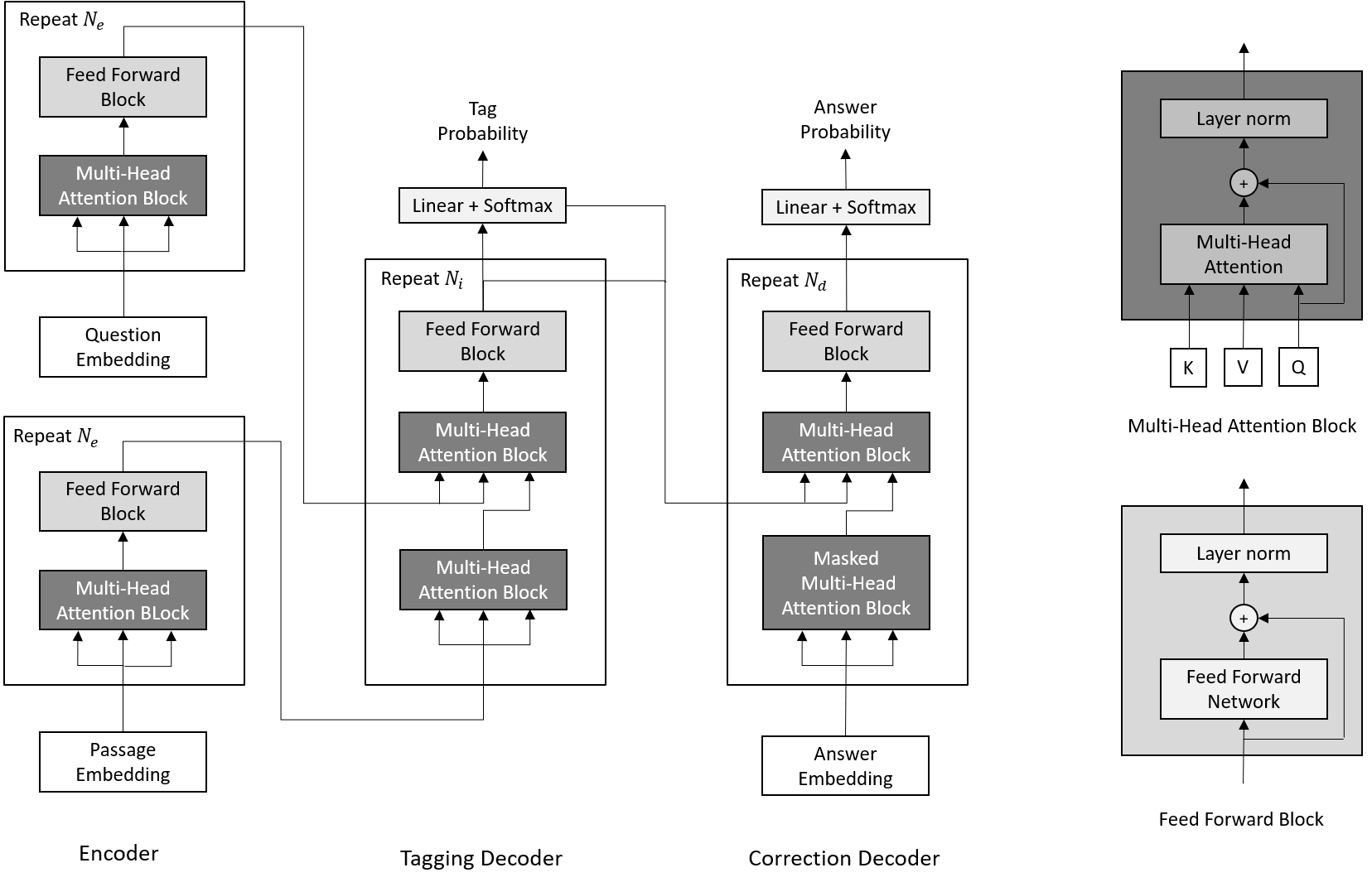}

\caption{Overview of two-stage model. Multi-Head Attention Block has three inputs, query(Q), key(K) and value(V). In this figure, we draw them in order of K, V and Q for clarity. For the answer embedding, we use entire ground truth answer in training. In decoding, the correction decoder generates answer word one by one. So we only use the generated answer word as input to generate next word.}
\label{figure_model}
\end{figure*}

Our model consists of three parts which contains an encoder, tagging decoder and correction decoder. We show our model in Figure \ref{figure_model}. We used two same-structure encoders to encode question and passage separately. Then the tagging decoder interacts between the encoded question and passage and tags each word in passage about its semantic role in the answer. The tagging decoder tags all passage words at same time. The correction decoder then generates an answer based on the tagging result. The correction decoder generate answer words one by one. Our intuition is to use tagging decoder to highlight the words in an article and use correction decoder to generate a fluent answer based on tagging results. As our approach's potentials, we show not only our approach being effective over this dataset but also the correction decoder in correcting the missing words which are not tagged as ground truth.

We use an example in Table \ref{table_tagging} to show our idea. The $argument_0$ is not a span of passage, the ''in'' is far from ''british columbia'', but most of the words in the answer are from the passage. The first stage of our model is to tag keywords from the passage. In this case, our tagging decoder tags all location words as argument such as ''vancouver'' and ''british columbida''. But the tagging result misses ''in''. The second stage is to generate a fluent answer based on the tagging result. In this case, our correction decoder adds ''in'' compared to the tagging result. We also could noticed that model also able to remove positive adverb "primarily". Based on our case study, the correct model is good at word ordering and small post-editing guided by language model.

\subsection{Encoder}
The encoder of our model contains a question encoder and a passage encoder, which is been used to encode the question and passage separately. These two encoders have the same structure but different weights in implementation.
The encoder is composed of two  basic  building  blocks, Multi-Head Attention Block and Feed Forward Block~\cite{transformer}. We will introduce these two building blocks and how to build an encoder with them.

\subsubsection{Multi-Head Attention Block (MHBlock)}
The core layer of the Multi-Head Attention Block is the Multi-Head Attention Layer ~\cite{transformer}. The input of Multi-Head Attention Layer contains query($Q$), key($K$) and value($V$). All the inputs are matrices. $Q\in\mathbb{R}^{n_q\times  d_k}$, $K\in\mathbb{R}^{n_k\times d_k}$, $V\in\mathbb{R}^{n_k\times d_v}$. The output $O$ of Multi-Head Attention Layer is a matrix too. $O\in\mathbb{R}^{n_q\times d_v}$. We represent this layer as a function $MultiHeadAttention(Q, K, V)$.

Intuitively, this layer is a soft dictionary lookup layer in vector space, and all the operation unit is vector. The dictionary in computer science is a set of key value pairs, lookup in dictionary is to find the key which equals to query and return corresponding value as output. In Multi-Head Attention, there are $n_k$ key value pairs, each key is a vector with dimension $d_k$ and each value is a vector with dimension $d_v$. The $n_q$ queries will have $n_q$ corresponding output. For each query, we will calculate attention score to each key, and use attention score as weight to calculate the weighted sum of value. The weighted sum is the output. More details are at Vaswani et al. (2017). 

Multi-Head Attention Block has same input as Multi-Head Attention Layer. But it requires the $d_k=d_v$. The inputs will go through a Multi-Head Attention layer wrapped with residual connection. Then pass the output though a layer norm layer to get the final output.\\

\begin{equation}
\begin{split}
MHBlock(Q,K,V)=& \\
LayerNorm(Q+&MultiHeadAttention(Q,K,V))
\end{split}
\nonumber
\end{equation}
\subsubsection{Feed-Forward Network Block (FFNBlock)} The core layer of Feed Forward Block is Feed-Forward Network ~\cite{transformer}. Feed-Forward Network is a two-layer projection on each row of matrix.
\begin{equation}
FFN(x)=max(0, xW_1+b_1)W_2+b_2 \nonumber
\end{equation}

The Feed-Forward Block has same input and output with Feed-Forward Network. We add the input and the output of Feed-Forward Network, then pass through a layer norm layer to get the final output.

\begin{equation}
\begin{split}
FFNBlock(x)=LayerNorm(x+FFN(x))
\end{split}
\nonumber
\end{equation}

\subsubsection{Encoder Structure}
An encoder is used to map a sequence of words into a sequence of hidden states. The question encoder and the passage encoder have the same structure. The input of the encoder is the word embedding of each word plus the position embedding. We use sine and cosine position embedding
~\cite{transformer}. The encoder is composed of a stack of $N_e$ identical layers. The output of the question encoder and the passage encoder are $h_q$ and $h_p$ respectively. For the question encoder:

\begin{equation}
\begin{split}
h_{q,0}&= Embedding(Q)+W_{pos}\\
h_{q,i}^m&=MHBlock(h_{q,i-1},h_{q,i-1},h_{q,i-1})\quad\forall i\in[1,N_e]\\
h_{q,i}&=FFNBlock(h_{q,i}^m)\quad\quad\quad\quad\quad\quad\quad\;\; \forall i\in[1,N_e]\\
h_q&=h_{q,n_e}
\nonumber
\end{split}
\end{equation}
$Embedding(x)$ is an embedding look up function, It takes the word id and output corresponding word embedding vector. $W_p$ is position embedding. $h_{q,i}^m$ is an intermediate result. Passage encoder has same structure, so we don't formulate it again.

\subsection{Tagging Decoder}
A tagging decoder is used to generate the tagging probability distribution for each word in a passage given a question encode result  $h_q$ and a passage encode result $h_p$.
In this sub subsection, we will introduce the tag format Semantic BIO Tags, and tagging decoder structure.
The output of the tagging decoder is a distribution of tags $T$. Formally, for each word $p_i$ in passage, the tagging decoder outputs a distribution $p(t_i|P,Q)$. $t_i$ is the tags for i-th passage word. We denote the result as $T=\{p(t_1|P,Q),p(t_2|P,Q),...,p(t_m|P,Q)\}$. In our model, we keep it as a continuous probability distribution so as to back propagate loss. If we want to give each word an explicit tag, we output the tag with maximum probability. 

\subsubsection{Semantic BIO Tags} 
We use semantic BIO tag like Stanovsky et al. (2018) to tag passage word. Each tag is combined by two parts, semantic tag for semantic role in answer and BIO tag for position in a field. The semantic role tag contains subject(S), predicate(P) and arguments(A). Since there are more than one argument, arguments also been distinguished by position as Ai for i-th argument. BIO tag contains Begin(B), Inside(I) and Outside(O). For each continuous subsequence belongs to same semantic role, we tag the first word to B, and tag the rest words to I. After tagging all continuous subsequence, we tag all the otherwords too. Then we add semantic role to BIO tags. If the semantic role is predicate,the tag will be extended to P-B, P-I. We showed an example at Table 2. The predicate has two words ''was filmed'' but they are not consecutive. For each sub span, we tag the first word as P-B. Then both the tag of ''was'' and ''filmed'' are P-B.So the re maybe more than one word been tagged as P-B although there is only one predicate in answer. For the same reason, there are two A1-B for ''in'' and ''british'' in example.

\subsubsection{Tagging Decoder ground truth} 
We need to create ground truth for tagging decoder training by ourself because the answer in dataset is in tuple format. Intuitively, when answer tuple was been created, some words were been selected from passage and copied to answer. Ideally, we want to tag these words out and let our model generate answer based on them too. Formally, we need to select out some continuous subsequences based on answer and tag them by previous tagging rules. Each subsequence must belong to same semantic role, but each semantic role may correspond to several subsequences.
The key challenge of it is that one word may have multiple occurrences in passage. We proposed a rule-based solution for this problem. Intuitively, for adjacent words in answer, we prefer to match the adjacent words in passage. For each field of answer, we prefer to keep all matches as close as possible. For details, we match the fields in the order of arguments, subject and predicate, because arguments is longest, and predicate is shortest. Then for each field, we try to match all bi-gram in passage. We will keep the multiple occurrence if exist. Then we match as much as possible single word which haven't covered by matched bi-gram. In this step, we will minimize the distance between rightmost word to leftmost word. In Open IE task, the predicate is shortest and often is unigram. So we match it at last and we prefer the predicate between subject and arguments.

\subsubsection{Tagging Decoder Structure}
Compared to the question encoder, the tagging decoder also needs to encode the passage. The difference is it needs to interactive with the question. We achieve this requirement by adding an additional attention layer from passage to question. The tagging decoder is composed of a stack of $N_t$ identical layers. Each layer consists of three sub-layers, self attention layer, passage to query encoding layer and feed forward layer. Self-attention layer is a Multi-Head Attention Block used to encode passage. The query, key and value of it is identical and is the output of previous layer. The passage to question layer is a Multi-Head Attention Block which is used to interactive with question. It's query is the output of previous self-attention layer, the key and value is identical and is the output of question encoder $h_q$. We also tried interactive layer like BIDAF (Seo et al. 2016), but there is no improvement compared to our model. Formally:

\begin{equation}
\begin{split}
h_{t,0}&= h_{p}\\
h_{t,i}^{t_0}&=MHBlock(h_{t,i-1},h_{t,i-1},h_{t,i-1})\quad \forall i\in[1,N_t]\\
h_{t,i}^{t_1}&=MHBlock(h_{t,i}^{t_0},h_{q},h_{q})\quad\quad\quad\quad\quad \forall i\in[1,N_t]\\
h_{t,i}&=FFNBlock(h_{t,i}^{t_1})\quad\quad\quad\quad\quad\quad\quad\;\: \forall i\in[1,N_t]\\
h_{t}&=h_{t,N_t}\\
\nonumber
\end{split}
\end{equation}
$h_{t,i}$ is the output of i-th layer, $h_{t,i}^{t_0}$ and $h_{t,i}^{t_1}$ is two intermediate results in same layer. $h_{t}$ is the final output of these $N_t$ layers.

Then we used linear projection and softmax on each item of $h_t$ to calculate the tag probability distribution $p(t_i|P,Q)$ of each passage word $p_i$.
\begin{equation}
p(t_i|P,Q)=softmax(h_{t,i}*W_t) \nonumber
\end{equation}
$h_{t,i}$ is the i-th row of $h_t$. $W_t$ is a linear projection matrix.

We use a semantic BIO tag like Stanovsky et al. 1. ~\cite{stanovsky2018supOpenIE} to tag passage words. Since there are multiple arguments, arguments are also distinguished by position as $A_i$ for the i-the argument. 
We need to create ground truth for tagging decoder training on our own because the answer in the dataset is in tuple format. Therefore we tag answer words which is in the passage and tag them as close as possible.
\subsection{Correction Decoder}
The correction decoder takes the output of the tagging decoder $h_{t}$ and $T$ as input, and generate a new answer. The correction decoder will generate answer words one by one like machine translation. 

We concatenate the answer tuple to one string as the output of the correction decoder. Formally, we concatenate the tuple to a sequence of l words $A=\{a_1,a_2,...,a_l\}$ which is formatted as "$subject$ \textless split\textgreater\ $predicate$ \textless split\textgreater\ $argument_1$ \textless split\textgreater\ ...\textless split\textgreater\ $argument_k$". The "\textless split\textgreater" is an additional format word used to identify the semantic role. We use "\textless split\textgreater" to separate multiple tuples for structured answer representation. We concatenate them into one string by "\textless split\textgreater" tag as our string version answer. We can choose a structured version or a string version as our output. The only difference between a structured version and a string version is whether it has <split> tag.

The structure correction decoder is very similar to tagging decoder. it also is composed of a stack of $N_c$ identical layers. Each layer consists of same three sub-layers, except first layer is a masked self-attention layer. The input is the sum of answer word embedding and position embedding too. 
Different to tagging decoder, the first layer, masked self attention layer, contains additional memory mask. This memory mask only allows the hidden vector at position i to pay attention on hidden vector before position i. This is because the generative decoder is a step by step decoder.we only have hidden vector before position i when we generate i-th word. 
Answer to passage encoding layer also is a Multi-Head Attention layer. The query of it is the output of masked self attention layer. the key and value of it is identical and is concatenation of two parts, tagging decoder hidden state $h_t$ and tagging result $T$.

In training, we could train the answer word decoding parallel by masked attention tricks. The only structure's difference between tagging decoder and correction decoder is that word in correction decoder only could attend to previous words. Because in decoding, we must generate answer word one by one. Suppose we had generated the first j-1 words: 
\begin{equation}
    \begin{split}
        h_{c,0}&=Embedding(concat(BOS,a_{<j})+W_{pos} \\
        h_m&=concat(h_t,T)\\
        h_{c,i}^{c_0}&=MHBlock(h_{c,i-1},h_{c,i-1},h_{c,i-1})\quad \forall i\in[1,N_c]\\
        h_{c,i}^{c_1}&=MHBlock(h_{c,i}^{c_0},h_m,h_m)\quad\quad\quad\quad\;\: \forall i\in[1,N_c]\\
        h_{c,i}&=FFNBlock(h_{c,i}^{c_1})\quad\quad\quad\quad\quad\quad\quad\;\: \forall i\in[1,N_c]\\
        h_{c}&=h_{c,N_c}
    \nonumber
    \end{split}
\end{equation}
$BOS$ is a special word at the beginning of a sentence. $h_{c,i}$ is the output of i-th layer, $h_{c,i}^{c_0}$ and $h_{c,i}^{c_1}$ are two intermediate results in the same layer. 

Then we generate j-th word by:
\begin{equation}
    \begin{split}
        p(a_j|a_{<j},P,Q)&=softmax(h_{c,j}*W_c)\\
        a_j&=argmax(p(a_j|a_{<j},P,Q))
    \end{split}
    \nonumber
\end{equation}
$h_{c,j}$ is the j-th row of $h_c$. $W_c$ is a linear projection matrix.

\subsection{Training}
In training, we create a ground truth for the tagging decoder and the correction decoder according to the previous method. These two decoders have separate loss but we train them jointly.

The loss of the tagging decoder is the negative log likelihood of ground truth tags.
\begin{equation}L_{tag}=-\frac{1}{N}\sum_{<P,Q>}\sum_{i=1}^mlogp(t_i|P,Q)\nonumber \end{equation}
$t_i$ is the ground truth tag of i-th word in passage. $N$ is the number of samples.\\

For generation, the loss is conditionally negative log probability of the ground truth answer.
\begin{equation}L_{correct}=-\frac{1}{N}\sum_{<P,Q,A>}\sum_{i=1}^{l}logp(a_i|a_{<i},P,Q) \nonumber \end{equation}
$a_i$ is the i-th word of answer. $N$ is the number of samples.

The loss of our model is the weighted sum of tagging decoder loss and correction decoder loss.
\begin{equation}L=\lambda*L_{tag}+L_{correct}\nonumber \end{equation}

$\lambda$ is the weight of $L_{tag}$ and will be tuned on validation set.

\section{Experiments}

\begin{table*}[t!]
	\small
    \begin{center}
    	\begin{tabular}{l|c|c|c|c}
        	\hline \bf Model & \bf Answer (BLEU-4) & \bf Subject (BLEU1) & \bf Predicate (BLEU1) & \bf Arguments (BLEU1) \\
            \hline
            Seq2Seq + Attention ~\cite{yan2018assertion} & 31.85 & - & - & - \\
            Seq2Ast ~\cite{yan2018assertion} & 35.76 & - & - & - \\
            Tagging & 55.60 & 51.61 & 57.19 & 46.62 \\
            Tagging + Correction & 59.32 & 63.40 & 67.50 & 61.01 \\
            \hline
            w/o question & 56.71 & 63.02 & 64.03 & 56.89 \\
            w/o semantic tag (Only BIO tag) & 58.78 & 62.61 & 66.36 & 59.72 \\
        	\hline
        \end{tabular}
    \end{center}
    \caption{\label{main-result} Test results on WebAssertions.}
\end{table*}

\subsection{Dataset and Evaluation}
\subsubsection{Dataset} We use the WebAssertions dataset  (Yan et al. 2018) to evaluate our model. The WebAssertions is a Question aware Open IE dataset. To construct this dataset,  Yan et al. (2018) collect queries from search engine logs as questions, retrieving and filtering related passages by search engine which cab directly answer the question. Then they extract answer tuples from the passage with Open IE model ClausIE. The labeler will judges whether the answer tuple has complete meaning and can answer the question. The answer tuple that has a positive label is the final answer. About 40\% of answers contain a field which is not a span of the passage. For example, sometimes the answer will delete words in a passage. Some words in the correction ground truth don't appear in the passage, and they thus don't appear in the span that tags result. This dataset contains 358,427 (question, passage, answer) triples. We randomly split the WebAssertions dataset into training, validation, and test sets with 8:1:1 split. We use the validation set to tune the model and report the results on the test set.

\subsubsection{Evaluation} We evaluate the quality of the entire answer and each semantic role. For entire answer, we concatenate the answer tuple to a string and split different role with the special word "\textless split\textgreater". Since there is only one subject and one predicate in answer, we can evaluate them directly. But there may be more than one argument, we concatenate them to one string by "\textless split\textgreater", just like the entire answer.

We use BLEU-4 score ~\cite{papineni2002bleu} as the evaluation metric for the entire answer, so as to be comparable with previous work. The subject and predicate are relatively short with an average length of 3.3 and 1.4. Therefore, we used BLEU-1 to evaluate each semantic role.

\subsection{Implementation Details}
\subsubsection{Data Processing}
We need a post process the output to get subject, predicate, and arguments separately. For tagging results, we collect all the words with same semantic tag to produce the corresponding answer field. The selected words are concatenated according to their order in passage. If no word is been tagged as one semantic role, then the result is an empty string. For the generated answer, we split them to a phrases list by the special split word. Then the first phrase is the subject, the second phrase is the predicate, and all other phrases are arguments.
We use byte-pair encoding (BPE)~\cite{philigage1994bpe} to handle the out of vocabulary problem in the correction decoder. BPE will split each word to several subwords. We control the corpus distinct subwords number. In the ground truth creation of semantic BIO tags, we match the continuous subsequence at the word level, and map the semantic tag to sub-word level and label BIO tag at subword level. After the model output tagging result, we collect the subword belonging to the same semantic role and undo BPE. We ignore the possibility of incomplete words and just let the model learn. For the generation, we split output to phrases list first and undo BPE on each phrase.

\subsubsection{Setting} We tune our hyper-parameter on the  validation dataset. The hidden size of our model is 512. We use shared vocabulary between question, passage, and answer. The BPE vocabulary size is 37000. We share the embedding weight for the question encoder, passage encoder, correction decoder, and pre-softmax linear transformation of the correction decoder. We use 8 head for Multi-Head Attention. The question and passage encoder layer number Ne is 2, the tagging decoder layer number $N_e$ Nt is 4. The correction decoder layer number $N_t$ is 4. The correction decoder layer number $N_c$ is 6. The weight of loss $\lambda$ is set to 3. We use ADAM optimizer ~\cite{kingma2014adam} to update model parameters and set the learning rate to 0.001.

\subsection{Baseline} Our proposed model is called \textbf{Tagging + Generation}. We compared it with three baselines.
\begin{itemize}
	\item
    {\bf Seq2Seq + Attention ~\cite{yan2018assertion}} This model formulates this task to a sequence to sequence problem. They concatenate question and passage to a string as input, and concatenate the tuple to a string as output. They insert special tag "\textless EOQ\textgreater" between question and passage, and special tag "," between field of tuple for format. This model uses a bidirectional GRU as encoder, GRU as decoder, and used attention mechanism.
	\item
    {\bf Seq2Ast ~\cite{yan2018assertion}} This sequence to assertion model (Seq2Ast) has the same input process and encoder as Seq2Seq + Attention. The difference is this model used a hierarchical decoder which first generates a representation for each field by a tuple-level decoder, then generates the words for each field by a word-level decoder.
    \item
    {\bf Tagging} We remove the correction decoder and only train the tagging decoder.
\end{itemize}

The Yan et al. ~\cite{yan2018assertion} also propose an extractive method, but it is not comparable with generative methods. This method extracts all possible answer tuples from the passage first. We use a ranking model to select the best answer as output. This dataset also is constructed by extracting tuples and their extraction model using the same extractor. The right answer is always in the ranking list. The key challenge with extractive methods is how to design the matching model. It is evaluated by ranking metrics, such as MAP, MRR. If we evaluate it with BLEU, it will reach 72.27. This result is higher that our result, but it is also reasonable because they leverage the dataset construct property. Our method does not rely on any Open IE model, so it still does not work well on dataset constructed in this way.

\subsection{Experiment Result}
The results are in Table~\ref{main-result}. Both the result of the entire answer and each semantic role show the same trend. We see the following: (i) the Tagging + Correction model achieves the best results which proves the effectiveness of our model; (ii) the Tagging+ Correction model is better than the Seq2Seq model. It means that by tagging the keyword first improves generation quality, which we think is because the tagging decoder provides a guide for the correction decoder; (iii) the Tagging + Correction model is also better than the tagging decoder, which means the second step correction is necessary.

For the subject, predicate, and arguments column, we find the results for the predicate are better than for subject, and subject results are better than arguments. This may be because of the different properties of different semantic roles. The subject is often a noun phrase. The predicate is a verb and has an average length of 1.4. The arguments are modifying phrase, which are longest and most complicated. Intuitively, the property of one word is enough to determine whether it is a predicate. The property of two adjacent words is enough to determine the boundary of a noun phrase. But we may need more sophisticated sentence information to extract arguments like syntax tree. We will leave this as future work to improve our model.

We remove the question, so it becomes an Open IE problem. We denote it as  \textbf{w/o question}. The entire answer result on BLEU will decrease by 2.6 compared with Tagging + Correction, and all the semantic role result will decrease too. This proves the Question aware Open IE cannot be solved as an Open IE task.

We also try to remove the semantic role in tags and only keeps the BIO tags. The results \textbf{w/o semantic tag} show that the BLEU of the entire answer will decrease 0.56, and BLEU of each semantic role also will decrease more than 1. This proves the semantic tag benefits from the correction decoder.

\subsection{Case Study}

We also do a case study to analysis our result. We randomly sample 50 samples in test dataset and predict with Tagging
+ Correction model. The summarization of the results is in Table 
 ~\ref{table-case-study}.

\begin{table}[h]
\centering
\begin{tabular}{l|c}
\hline \bf Label & \bf Ratio \\ \hline
correct / exactly match & 30\% \\
correct / comparable & 10\% \\
correct / better & 18\% \\
correct / incomplete label & 18\% \\
\hline 
wrong / wrong focus & 12\% \\
wrong / grammar problem & 6\% \\
wrong / lost key words & 6\% \\
\hline
\end{tabular}
\caption{Case study of Tagging + Correction.}
\label{table-case-study}
\end{table}

We find that about 76\% of cases are correct. \textbf{Comparable} means the model output is comparable with ground truth and it is hard to tell which one is better. About 18\% of cases are \textbf{better} than the ground truth. This is because the generated answer is shorter and clearer than the ground truth answer, especially on arguments. Another 18\% of wrong cases are because of an \textbf{incomplete label}, which means there are more than one answer in the passage for the question. Based on these results, we see that it is hard to evaluate the Question aware Open IE because of the open definition of information extraction problem. There may be more than one answer in the passage and each answer may have multiple paraphrases. A better dataset could help to solve the ''better'' and ''incomplete label'' problems.

For the wrong output, about 12\% of wrong cases are because of the  \textbf{wrong focus}. This means the answer is not related to the question. 6\% of cases are because of a  \textbf{grammar problem}, which means the answer is not fluent. This is because the language model of the correction decoder is still not good enough. 6\% of cases are because of lost key words. In the future, we could try to improve the interaction between question and passage to improve the wrong focus and lost key words problem. In the future, we may try to improve the interaction between question and passage to improve the wrong focus and lost key words problem. We could also try to transfer learning to improve the language model.

\section{Conclusion}
In this paper, we introduce a two-stage decoder model to solve the question aware Open IE task. Because most of the answer words are from a passage, we use a tagging decoder to tag the key words in the passage first, and generate a refined answer with a correction decoder based on the output of the tagging decoder. The experiments on WebAssertions show that our method outperforms other pure generation models or tagging models. Our model does not rely on any Open IE tools which gives it good generalization ability. In the future, we will try more methods to improve our results like incorporate syntax information or more interaction methods. We will also consider creating a better dataset to accelerate research in this area.




\end{document}